\documentclass[journal]{IEEEtran}
\usepackage{cite}
\usepackage{amsmath}
\usepackage{amsfonts}
\usepackage{pifont} % 打勾 and 打叉
\usepackage{graphicx}
\usepackage{float}
\usepackage{subfigure}
\usepackage{rotfloat} % 单元格旋转
\usepackage{colortbl} % 彩色表格
\usepackage{color}
\usepackage[svgnames]{xcolor}

\usepackage{tabu}% 表格插入
\usepackage{multirow}% 一般用以设计表格，将所行合并
\usepackage{multicol}% 合并多列
\usepackage{multirow}% 合并多行
\usepackage{makecell}% 三线表-竖线
\usepackage{booktabs}% 三线表-短细横线

\usepackage[colorlinks,
            linkcolor=blue,
            anchorcolor=green,
            citecolor=red,]{hyperref}

\begin{document}

\title{Multi-scale Restoration of Missing Data in\\%
        Optical Time-series Images with Masked\\%
        Spatial-Temporal Attention Network%
}

\author{
    Zaiyan~Zhang,~\IEEEmembership{Student~Member,~IEEE,}
    Jining Yan,~\IEEEmembership{Senior~Member,~IEEE,}
    Yuanqi~Liang,~
    \\Jiaxin~Feng,~
    Haixu~He,~
    and Li Cao

    \thanks{
        \emph{(Corresponding author: Jining~Yan.)}
    }
    \thanks{
        Z.~Zhang, J.~Yan, Y.~Liang, J.~Feng and H.~He are with the School of Computer Science, China University of Geosciences, Wuhan 430074, China, and also with the Engineering Research Center of Natural Resource Information Management and Digital Twin Engineering Software, Ministry of Education, Wuhan 430074, China (e-mail: zzaiyan@cug.edu.cn; yanjn@cug.edu.cn; naji@cug.edu.cn; fjxpen1234@cug.edu.cn; 20161001925@cug.edu.cn).
    }
    \thanks{
        L.~Cao is with the Second Surveying and Mapping Institute of Hunan Province, Changsha 410029, China, and also with the Key Laboratory of Natural Resources Monitoring and Supervision in Southern Hilly Region, Ministry of Natural Resources, Changsha 410029, China (e-mail: cl@img.net).}
}

% The paper headers
%\markboth{The manuscript submitted to IEEE Transactions of Geoscience and Remote Sensing}%
%{Zhang \textit{et al},: Multi-scale Restoration of Missing Data in Optical Time-series Images with Masked Spatial-Temporal Attention Network}

\maketitle

\begin{abstract}

Remote sensing images often suffer from substantial data loss due to factors such as thick cloud cover and sensor limitations. Existing methods for imputing missing values in remote sensing images fail to fully exploit spatiotemporal auxiliary information, which restricts the accuracy of their reconstructions. To address this issue, this paper proposes a novel deep learning-based approach called MS$^2$TAN (\underline{M}ulti-\underline{S}cale \underline{M}asked \underline{S}patial-\underline{T}emporal \underline{A}ttention \underline{N}etwork) for reconstructing time-series remote sensing images. First, we introduce an efficient spatiotemporal feature extractor based on Masked Spatial-Temporal Attention (MSTA) to capture high-quality representations of spatiotemporal neighborhood features surrounding missing regions while significantly reducing the computational complexity of the attention mechanism. Second, a Multi-Scale Restoration Network composed of MSTA-based Feature Extractors is designed to progressively refine missing values by exploring spatiotemporal neighborhood features at different scales. Third, we propose a “Pixel-Structure-Perception” Multi-Objective Joint Optimization method to enhance the visual quality of the reconstructed results from multiple perspectives and to preserve more texture structures. Finally, quantitative experimental results under multi-temporal inputs on two public datasets demonstrate that the proposed method outperforms competitive approaches, achieving a 9.76\%/9.30\% reduction in Mean Absolute Error (MAE) and a 0.56 dB/0.62 dB increase in Peak Signal-to-Noise Ratio (PSNR), along with stronger texture and structural consistency. Ablation experiments further validate the contribution of the core innovations to imputation accuracy.

\end{abstract}


\IEEEpeerreviewmaketitle

\section{Introduction}

\IEEEPARstart{O}{ver} the past few decades, remote sensing data has been extensively used in various industries. Among them, high spatial resolution remote sensing imagery is particularly beneficial for applications such as vegetation monitoring, land cover mapping, and land cover change detection. However, the fine spatial resolution images suffer from inevitable information loss caused by internal factors (e.g., sensor malfunction) and external factors (e.g., atmospheric conditions), which restrict their applications in different domains \cite{shen2015missing}.

This paper addresses the issue of missing data in remote sensing image involving multiple spectra. Common tasks include resolving the Landsat-7 ETM+ sensor scan line corrector off (SLC-off) problem and removing thick clouds. The key challenge is to estimate the missing regions and fill the gaps with predicted pixels, ensuring visual and semantic consistency with the surrounding pixels to enhance data usability.

Researchers have proposed various methods to recover missing data in remote sensing images. Early approaches to missing value restoration can be broadly categorized into three types \cite{shen2015missing}: spatial-based, temporal-based, and spatiotemporal-based data recovery. These methods have shown promising results in specific scenarios with low resolution and low missing rates. However, most of them rely on linear models and struggle to handle complex and detailed scenes. Moreover, due to limited reference information, the generated images often exhibit blurriness and lack continuous textures.

In recent years, the rapid advancement of deep learning theory and computer hardware \cite{lecun2015deep} has led to significant progress in remote sensing image restoration using deep learning-based methods. These methods can be broadly categorized into two types: CNN-based \cite{lecun1989handwritten} and ViT-based \cite{dosovitskiy2020image} approaches. Compared to traditional statistical models, CNNs exhibit strong non-linear expressive power, allowing for efficient extraction of spatial features from remote sensing data and significantly improving the accuracy of image restoration. However, these structures struggle to fully exploit time-series information, resulting in a bottleneck in restoration accuracy. ViTs, based on self-attention mechanism \cite{vaswani2017attention}, possess a global receptive field, enabling comprehensive and efficient utilization of both temporal sequences and images to enhance reconstruction results \cite{sts-fusion}. However, due to the high resolution and long time-series of remote sensing images, token sequences become excessively long, leading to extremely high complexity in attention computations.

To efficiently mine spatiotemporal information in remote sensing images, we apply self-attention mechanism separately in the temporal and spatial dimensions and alternate between the two \cite{bertasius2021space}, greatly reducing computational complexity. To address the problem of significant distributional differences between missing and non-missing values in remote sensing images, we apply missing values mask and diagonal mask to the attention matrix \cite{du2023saits}, proposing Masked Spatial-Temporal Attention (MSTA) to enhance the expressive power of spatiotemporal attention and optimize the spectral discrepancy at the transition regions.

For full exploitation of the spatiotemporal neighborhood features at different scales \cite{yang2017high}, we further propose a Multi-scale Restoration Network. The network consists of MSTA-based Feature Extractors with different embedding scales, which progressively refine the reconstruction of missing information from coarse to fine granularity levels, achieving higher restoration accuracy \cite{ResNet}. For model training, we propose a ``Pixel-Structure-Perception'' Multi-Objective Joint Optimization method, using pixel-wise loss as the basic loss for the restoration task and incorporating structural loss \cite{zhao2016loss} and perceptual loss \cite{johnson2016perceptual} to optimize the model's results from the perspectives of structure, texture, shape, and spatial relations, thereby achieving high-quality image inpainting.

Finally, we performed both quantitative and qualitative experiments on two datasets, comparing our approach with several mainstream methods. Additionally, we carried out ablation studies on the key innovations and analyzed the trade-offs between effectiveness and efficiency across models of different sizes.

In summary, our main contributions are as follows:

\begin{enumerate}
\item
We propose a deep learning-based method MS$^2$TAN for reconstructing missing data in time series remote sensing images. Our method utilizes a multi-scale restoration network to learn an end-to-end mapping between incomplete and complete image sequences. MS$^2$TAN achieved higher restoration accuracy than mainstream methods in quantitative experiments, and showed better visual effects in eliminating gaps in real data.
\item
To address the challenges of high resolution and long time series in remote sensing data, we introduce the Masked Spatial-Temporal Attention (MSTA) mechanism. MSTA effectively extracts spatiotemporal features, improves the utilization of spatiotemporal context information, reduces color transition artifacts at the boundaries of missing value areas, and significantly reduces the computational complexity of self-attention.
\item
For model training, we propose a ``Pixel-Structure-Perception'' Multi-Objective Joint Optimization method. This method considers pixel-wise reconstruction error, structural reconstruction error, and perceptual error, resulting in restoration results with enhanced visual quality and preserved texture and structural details.
\end{enumerate}

The rest of this paper is organized as follows. In Section \ref{RelatedWork}, we review the existing methods for reconstructing missing information in remote sensing images. The network architecture and methodology details of our proposed model are presented in Section \ref{Methodology}. In Section \ref{Experiments}, we showcase the results of missing data reconstruction in both quantitative and qualitative experiments, compare them with mainstream methods, and validate the effectiveness of each component through validation studies. Finally, our conclusions and future prospects are summarized in Section \ref{Conclusion}.

%\pagebreak

\section{Related Works} \label{RelatedWork}

\subsection{Traditional methods}

Early research on the restoration of missing data in remote sensing images can be roughly divided into three categories: spatial-based, temporal-based, and spatiotemporal hybrid methods.

\subsubsection{Spatial-based Methods}
Spatial-based methods rely solely on the valid information within the image itself to predict the missing data. The most commonly used approach is spatial interpolation methods \cite{gaps-fill}. Additionally, methods based on partial differential equations (PDE) \cite{Image-inpainting} and variational methods \cite{Hardie1997Variation, Yuan2014Variation, Cheng2014Variation} have also been utilized for reconstructing missing values. Furthermore, patch-based methods have found extensive application \cite{exemplar-based, HeKaiming2014, Cheng2017GlobalOptim}. In general, spatial-based methods are suitable for reconstructing small missing areas or regions with regular textures. However, the accuracy of the reconstruction cannot be guaranteed, particularly for large regions or complex textures.

\subsubsection{Temporal-based Methods}
Temporal-based methods utilize observations of the same location at different times from satellites to restore missing data. These methods include histogram matching-based approaches \cite{LLHM}, temporal interpolation-based methods \cite{NSPI}, replacement-based methods \cite{TSAM}, and regression-based methods \cite{zeng2013recovering, li2014recovering, zhang2014missing}, among others. However, the differences between different time phases restrict the application of these methods.

\subsubsection{Spatiotemporal Hybrid Methods}
To overcome these limitations, spatiotemporal hybrid methods integrate the spatial and temporal correlations to reconstruct missing data under various conditions. For instance, improved nearest neighbor pixel interpolation methods \cite{zhu2011modified}, methods based on spatiotemporal Markov random field model \cite{Cheng2014Patch}, spatiotemporal weighted regression model \cite{STWR}, methods based on group sparse representation \cite{li2016patch}, and methods based on low-rank tensor decomposition \cite{AWTC, HHX} have been proposed. However, most of these methods rely on linear models and struggle to handle complex and intricate scenes.

\subsection{Deep learning-based methods}

In recent years, deep learning methods based on CNN and Transformer have been widely applied.

\subsubsection{CNN-based methods}

CNNs have shown high efficiency in extracting spatial features from remote sensing data, leading to significant improvements in the accuracy of remote sensing image restoration. Malek et al. \cite{malek2017reconstructing} applied a context encoder \cite{context-encoder} to reconstruct thick clouds in remote sensing images. CNNs combined with GAN structures \cite{sun2019cloud, shao2022efficient} were used to generate realistic reconstructed images. Zhang et al. \cite{STS-CNN} proposed a CNN-based spatiotemporal Spectra (STS-CNN) framework, which was further developed into a progressive spatiotemporal patch grouping framework \cite{zhang2020thick}. CNNs that incorporate temporal inputs through channel-wise concatenation \cite{chen2019thick} merge feature maps of target images and temporal images, introducing auxiliary information for missing data restoration. Stucker et al. \cite{stucker2023u} used temporal self-attention for CNN's feature map sequences in conjunction with U-Net to repair temporal images. 
\textbf{However}, CNNs lack a true understanding of time-series, which hinders the efficient integration of temporal and spatial information in these methods and limits the utilization of long temporal sequences as auxiliary information, ultimately affecting the restoration accuracy.

\subsubsection{ViT-based methods}

Visual Transformer (ViT) \cite{dosovitskiy2020image} has shown excellent performance in many tasks in the vision domain. Xu et al. \cite{xu2022attention} applied spatial self-attention to feature maps to capture the distribution of cloud thickness. Christopoulos et al. \cite{christopoulos2022cloudtran} utilized axial attention to remove thick clouds in remote sensing images. Recently, Liu et al. \cite{liu2023thick} used spatial attention and channel attention to remove cloud cover in the images. 
\textbf{However}, the self-attention used in ViT requires computing the correlations between all pairs of patches (including all times). In the context of processing time-series remote sensing images, the images often have high resolutions and long time-series, resulting in a large number of tokens and extremely high computational complexity. 

Bertasius et al. \cite{bertasius2021space} conducted a detailed comparison of various forms of spatial-temporal attention and proposed separated spatial-temporal attention that achieves a balance between efficiency and performance. We further improved the separated spatial-temporal attention by introducing missing value masks and diagonal masks \cite{du2023saits}, resulting in Masked Spatial-Temporal Attention (MSTA), which exhibits superior performance in the task of missing value restoration. Compared to CNN-based methods and original ViT methods, MSTA enables efficient processing of spatiotemporal information and leverages long-term temporal information to assist in missing value reconstruction, leading to more precise reconstruction results.

\begin{figure*}[t]
\centering
\includegraphics[width=.95\linewidth]{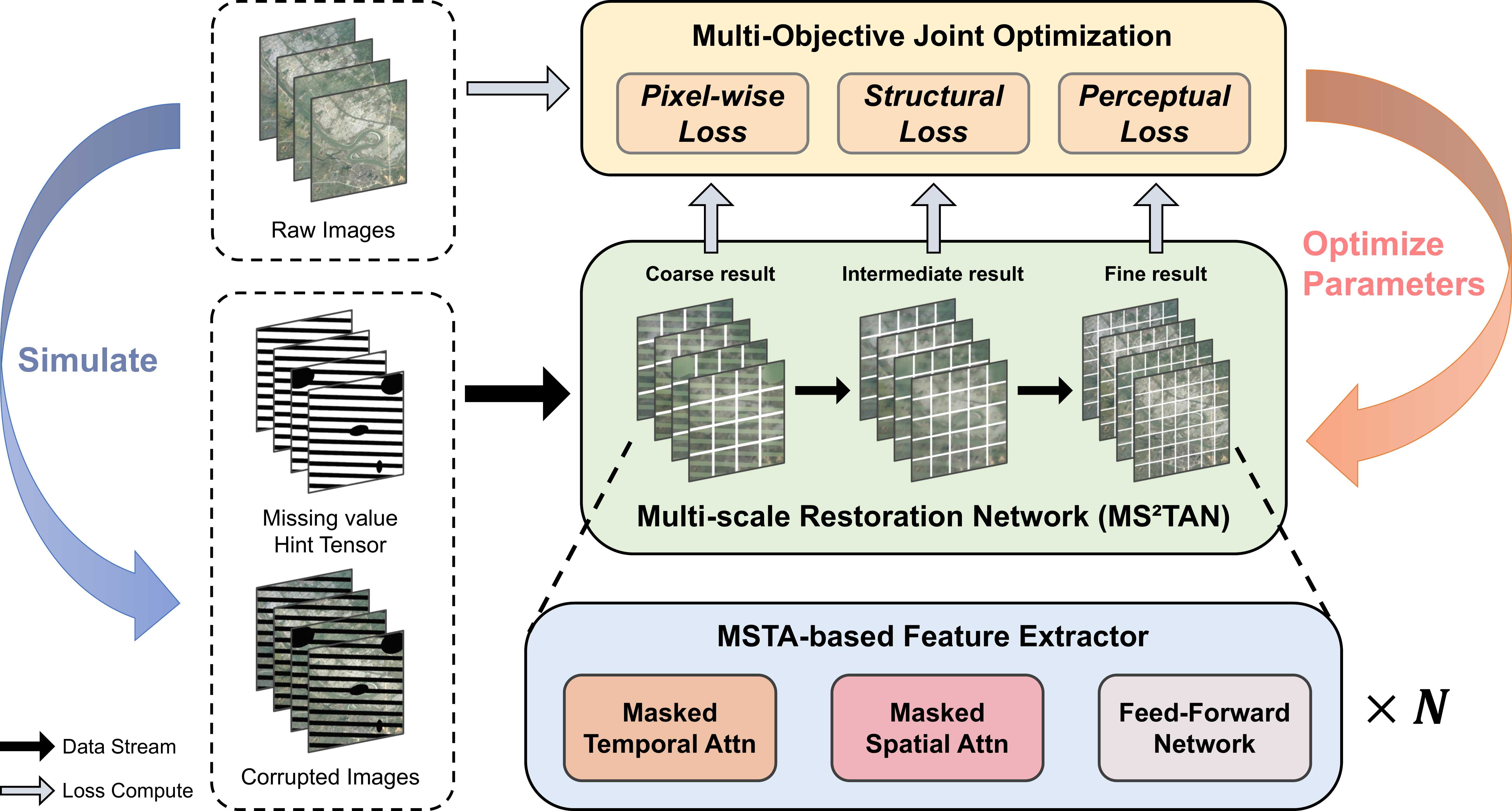}
\caption{The overall flowchart of the proposed method consists of two main components: a Multi-scale Restoration Network (named MS$^2$TAN) with MSTA-based Feature Extractors, and a ``Pixel-Structure-Perception'' Multi-Objective Joint Optimization method.}
\label{fig:framework}
\end{figure*}

\section{Methodology} \label{Methodology}

\subsection{Problem Definition and Overall Framework}

\subsubsection{Problem Definition}

The original time-series remote sensing image with missing values can be represented as \(X \in \mathbb{R}^{T \times C \times H \times W}\), where $T$ denotes the length of the time-series, $C$ represents the number of channels, and $H$ and $W$ denote the spatial dimensions of the region. $X_{(t, c, i, j)}$ denotes the value of channel $c$ at position $(i, j)$ at time $t$. To account for the missing values, we introduce the missing value hint tensor $M \in \mathbb{R}^{T \times C \times H \times W}$, which is defined as shown in Eq. \eqref{MissingMask}. In practical applications, the missing value hint tensor $M$ can be obtained from QA band or cloud detection algorithms such as Fmask \cite{zhu2012object} and S2Cloudless \cite{baetens2019validation}.
\begin{equation}
M_{(t, c, i, j)} = 
\left\{
\begin{matrix} 
  0 & \text{if}~X_{(t, c, i, j)}~\text{is~missing},\\  
  1 & \text{if}~X_{(t, c, i, j)}~\text{is~observed}.
\end{matrix} \label{MissingMask}
\right.
\end{equation}

The target sequence \(Y\) in \(\mathbb{R}^{T \times C \times H \times W}\) represents the actual data, while \(\widetilde{Y}\) in \(\mathbb{R}^{T \times C \times H \times W}\) represents the data repaired by the model. Therefore, the definition of the multi-temporal remote sensing image reconstruction model is given by equation \eqref{Model}. Here, the output \(\widetilde{Y}\) of the model is the initial reconstruction result, and \(\odot\) denotes the Hadamard product. By replacing the observed real values in \(X\) with \(\widetilde{Y}\), we obtain the final reconstruction result \(\widetilde{Y}_{\mathrm{out}}\) as shown in Eq. \eqref{replaceOut}.
%\begin{equation}
\begin{gather}
\widetilde{Y} = \operatorname{Model}(X, M)\label{Model}\\
\widetilde{Y}_{\mathrm{out}} = \widetilde{Y} \odot (1-M) + X \odot M \label{replaceOut}
\end{gather}
%\end{equation}

Our objective is to make \(\widetilde{Y}\) closely match the data distribution of \(Y\) in order to achieve high-quality reconstructed images.

\subsubsection{Overall Framework}

The proposed framework for time-series image recovery consists of a Multi-scale Restoration Network (called MS$^2$TAN) with MSTA-based Feature Extractors and a ``Pixel-Structure-Perception'' Multi-Objective Joint Optimization method, as depicted in Fig. \ref{fig:framework}.

The framework utilizes the MS$^2$TAN to learn the non-linear mapping from non-missing information to missing information. It employs the MSTA-based Feature Extractor (MFE) at different scales to extract temporal and spatial features for reconstruction. Finally, the network parameters are optimized using the ``Pixel-Structure-Perception'' Multi-Objective Joint Optimization method, and the trained parameters are used for inference. The details of these components will be discussed in the following sections \ref{MFE}, \ref{MRN} and \ref{MOO}.

\begin{figure*}[t]
	\centering
	\includegraphics[width=.95\linewidth]{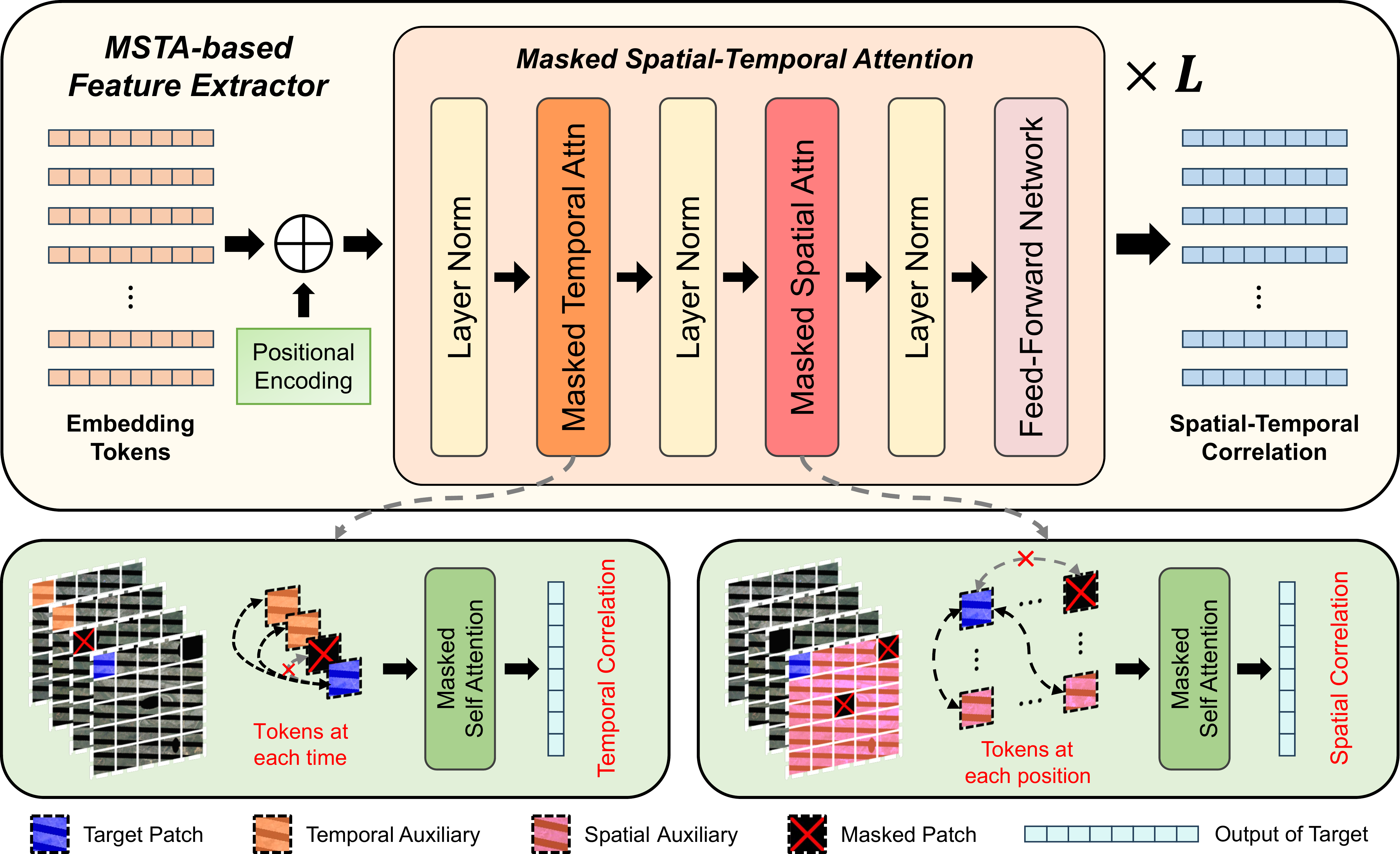}
	\caption{Illustration of the structure of the MSTA-based Feature Extractor (MFE). The input sequence is augmented with positional encoding to incorporate positional information. It then undergoes $L$ layers of Masked Spatial-Temporal Attention operations to obtain the output sequence, which combines temporal and spatial contextual features. The Masked Temporal/Spatial Attention, introduced below the image, respectively employ masked self-attention to capture the data distribution features in the Temporal/Spatial neighborhood.}
	\label{fig:MSTA}
\end{figure*}

\subsection{MSTA-based Feature Extractor} \label{MFE}

The structure of the MSTA-based Feature Extractor (MFE) module is illustrated in Fig. \ref{fig:MSTA}. It consists of position encoding and a cascade of $L$ Masked Spatial-Temporal Attention (MSTA) units. The input to this module is a high-dimensional feature vector $E \in \mathbb{R}^{TN \times d_{\mathrm{emb}}}$, which is obtained by embedding the time-series remote sensing image $X$ and the missing information mask $M$ into blocks (as described in detail in Section \ref{Emb-Unemb}). Here, $T$ represents the time-series length, $N$ represents the number of patches in a single remote sensing image, and $d_{\mathrm{emb}}$ represents the dimension of the token vector corresponding to each patch. Subsequently, the position encoding introduces spatial-temporal positional semantics, followed by $L$ layers of MSTA units, and ultimately produces a sequence with incorporated spatiotemporal correlation features.

\subsubsection{Masked Self-Attention}

The Transformer model utilizes self-attention mechanism to model sequences. It maps the input to query vector Q, key vector K, and value vector V. The attention score between Q and K is computed using Scaled Dot-Product, followed by the application of the Softmax function to obtain attention weights A. The final output is the attention-weighted V, as shown in Eq. \ref{SelfAttn}.
\begin{equation}
\begin{gathered}
\mathrm{H} = \mathrm{A} \mathrm{V}  = \mathrm{Softmax}\left(\frac{\mathrm{Q} \cdot \mathrm{K}^\top}{\sqrt{d_k}}\right) \cdot \mathrm{V}\\
\mathrm{where}\ \mathrm{[Q, K, V]} = X \cdot W_{\mathrm{QKV}}. 
\end{gathered} \label{SelfAttn}
\end{equation}

To address the challenge of disparate distribution of missing values in remote sensing data and enhance feature extraction capability, we introduce missing value masks and diagonal masks into self-attention. 

\begin{itemize}
\item The missing value mask sets the attention scores from patches with too high missing rate to other patches as $-\infty$, effectively masking the influence of missing values. This enables the model to focus on extracting useful information from non-missing data to fill in the gaps, without being affected by the missing data.
%\pagebreak
\item The diagonal mask sets the diagonal of the attention matrix as $-\infty$, preventing each step from attending to itself and forcing it to rely on the other $TN-1$ steps for inference. This helps capture the spatiotemporal feature correlation in high-dimensional space.
\end{itemize}

More specifically, for the attention score matrix $\mathrm{Sc} \in \mathbb{R}^{TN \times TN}$, the attention mask operation $\mathrm{ApplyMask}$ is defined as shown in Eq. \eqref{ApplyMask}.
\begin{equation}
\begin{gathered}
\left.\mathrm{ApplyMask}\left(\mathrm{Sc}\right)_{\left(i,j\right)}=\left\{\begin{array}{ll}
-\infty&\mathrm{Mask}(i, j)=1,\\
\mathrm{Sc}_{\left(i,j\right)}&\mathrm{Mask}(i, j)\ne1.
\end{array}\right.\right.\\
\mathrm{Mask}(i, j)=1\ \textbf{only if}\ (\mathrm{MR}(E(i))>C_\mathrm{max}\ \textbf{or}\ i=j)
\end{gathered} \label{ApplyMask}
\end{equation}

Here, $Mask()$ checks whether the mask condition is satisfied, $\mathrm{MR}(e)$ represents the missing rate of the corresponding patch $e$ (which can be calculated from the hint tensor $M$), and $C_\mathrm{max}$ is a hyperparameter that controls the maximum allowable missing rate. Applying Eq. \eqref{ApplyMask} to the attention scores in Eq. \eqref{SelfAttn}, we obtain the expression for a single masked attention head, as shown in Eq. \eqref{MaskedH}. Finally, by linearly projecting and combining multiple attention heads with the residual connection, we obtain the output of the masked self-attention, as shown in Eq. \eqref{MaskedAttn}.
\begin{gather}
\mathrm{H}^{(i)} = \mathrm{A}^{\prime} \mathrm{V}  = \mathrm{Softmax}\left(\mathrm{ApplyMask}\left(\frac{\mathrm{Q} \cdot \mathrm{K}^\top}{\sqrt{d_\text{qkv}}}\right)\right) \cdot \mathrm{V} \label{MaskedH}
\end{gather}
\begin{gather}
\mathrm{MaskedAttn}(X, M) = \mathrm{Proj}\left(\left[\mathrm{H}^{(1)}, \mathrm{H}^{(2)}, \cdots, \mathrm{H}^{(h)}\right]\right) + X \label{MaskedAttn}
\end{gather}

In Eq. \eqref{MaskedAttn}, $\mathrm{A}^{\prime}$ denotes the attention weights post mask application, $\mathrm{MaskedAttn}()$ represents the masked self-attention operation, $\mathrm{Proj}()$ stands for projection head, and $h$ denotes the number of attention heads.

\subsubsection{Masked Spatial-Temporal Attention}

Masked Spatial-Temporal Attention (MSTA) is used to capture the spatiotemporal correlations within the input sequence, and consists of Masked Temporal Attention (MTA), Masked Spatial Attention (MSA), Layer Normalization (LN), and Feed-Forward Network (FFN). These components are elaborated upon below.

For an input sequence $E \in \mathbb{R}^{TN \times d_{\mathrm{emb}}}$, MTA first reshapes $E$ to place the time dimension and the feature dimension at the end, obtaining a sequence $e_{t}$ of length $T$ and $d_{\mathrm{emb}}$ dimensions. Subsequently, for each position in space, MaskedAttn operation is performed along the temporal direction to obtain the output of MTA, as shown in Eq. \eqref{MTA}. 
\begin{equation}
\begin{gathered}
e_{t}=\mathrm{Reshape}(E) \in \mathbb{R}^{N \times T \times d_{\mathrm{emb}}}\\
\mathrm{MTA}(E, M)=\mathrm{Reshape}\left(\mathrm{MaskedAttn}\left(e_{t}, M\right)\right)
\end{gathered} \label{MTA}
\end{equation}

Similarly, after the corresponding dimension transformation of the input sequence, MaskedAttn operation is conducted along the spatial direction for each temporal image, yielding the output of MSA, as shown in Eq. \eqref{MSA}.
\begin{equation}
\begin{gathered}
e_{s}=\mathrm{Reshape}(E) \in \mathbb{R}^{T \times N \times d_{\mathrm{emb}}}\\
\mathrm{MSA}(E, M)=\mathrm{Reshape}\left(\mathrm{MaskedAttn}\left(e_{s}, M\right)\right)
\end{gathered} \label{MSA}
\end{equation}

FFN consists of two linear layers separated by a ReLU activation function, with a residual connection established between the input and output. The expression for FFN is presented in Eq. \eqref{FFN}.
\begin{gather}
\mathrm{FFN}\left(X\right)=\mathrm{Linear}\left(\mathrm{ReLU}\left(\mathrm{Linear}\left(X\right)\right)\right) + X \label{FFN}
\end{gather}

MSTA sequentially applies MTA and MSA to the input token sequence, exploring the spatiotemporal correlations among patches. Subsequently, FFN is used to fuse spatiotemporal features and introduce non-linear transformations, as shown in Eq. \eqref{MSTA}.
\begin{equation}
\begin{gathered}
U = \operatorname{MTA}(\operatorname{LN}(E), M)\\
V = \operatorname{MSA}(\operatorname{LN}(U), M)\\
\operatorname{MSTA}(E, M) = \operatorname{FFN}\left(\operatorname{LN}\left(V\right)\right)
\end{gathered} \label{MSTA}
\end{equation}

\begin{figure*}[t]
\centering
\includegraphics[width=\linewidth]{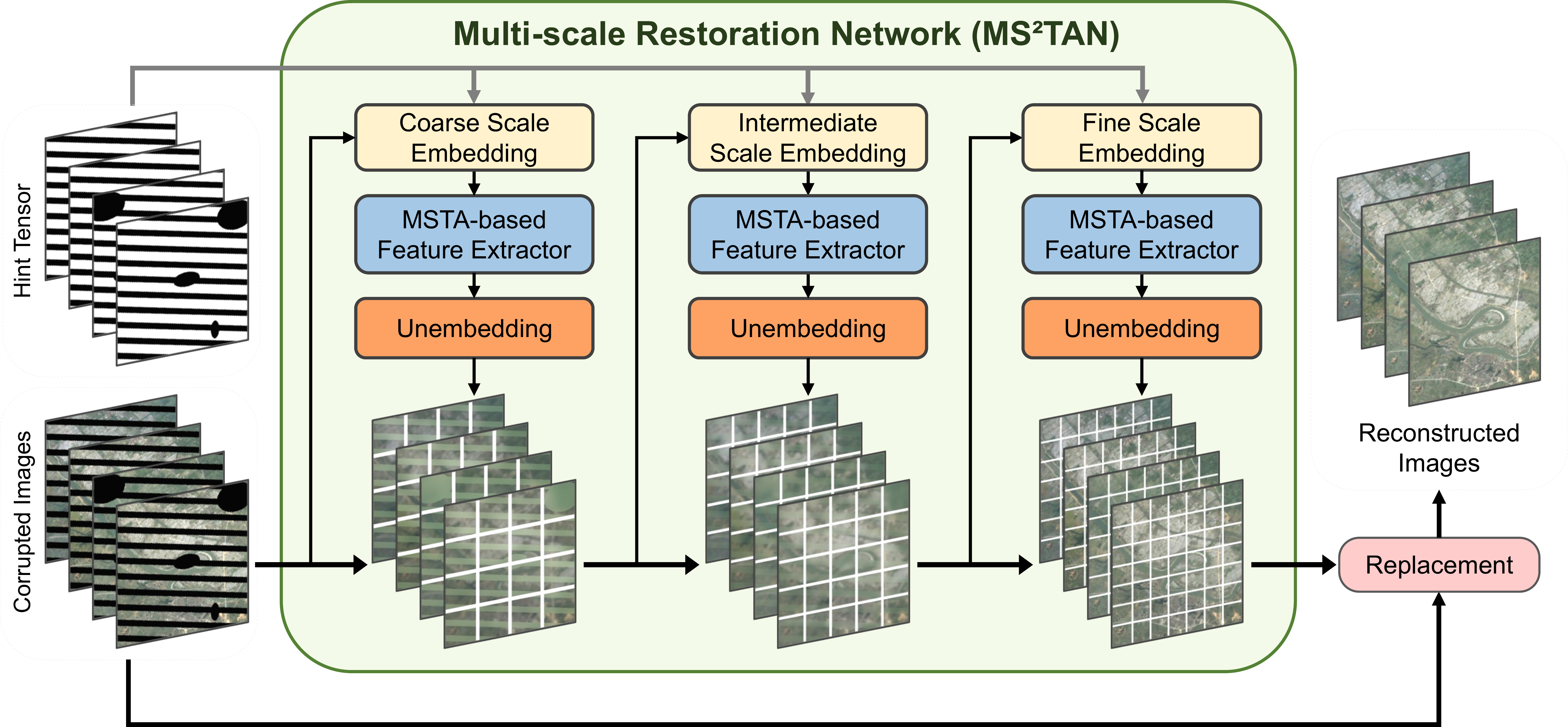}
\caption{The network structure diagram of proposed Multi-scale Restoration Network (MS$^2$TAN), which contains several residual-connected restoration modules consisting of Embedding, MFE, and Unembedding.}
\label{fig:MRN}
\end{figure*}

\subsubsection{Positional Encoding}

In the Transformer, positional encoding is added to the input sequences to introduce positional semantics, as depicted in Eq. \eqref{PosEnc}. Here, $pos = t \times N + n \in \left[0,~TN - 1\right]$ represents the spatiotemporal joint coordinate derived from the temporal index $t \in \left[0,~T - 1\right]$ and spatial index $n \in \left[0,~N - 1\right]$, where $N$ denotes the total number of patches in a single image. The function $\operatorname{PosEnc}(pos, dim)$ indicates the value of the $dim$-th dimension of the positional encoding for the $pos$-th patch in the image sequence.
\begin{equation}
\begin{gathered}
\operatorname{PosEnc}(pos, 2i)=\sin \left({pos} \cdot {10000^{-{2i}/{d_{\mathrm{emb}}}}}\right)\\
\quad \operatorname{PosEnc}(pos, 2i+1)=\cos \left({pos} \cdot {10000^{-{2i}/{d_{\mathrm{emb}}}}}\right)
\end{gathered} \label{PosEnc}
\end{equation}

After positional encoding, the input sequence will undergo $L$ layers of MSTA to explore deeper spatiotemporal features. Finally, the overall expression of the MFE module is shown in Eq. \eqref{MFE-Final}, where the symbol ${}^{L}$ denotes stacking $L$ layers, and $\mathrm{MFE}()$ represents the MSTA-based Feature Extractor.
\begin{equation}
\begin{gathered}
\mathrm{MFE}\left(E, M\right)=\{\mathrm{MSTA}\left(E+\mathrm{PosEnc} ,M\right)\}^{L}
\end{gathered} \label{MFE-Final}
\end{equation}

\subsection{Multi-scale Restoration Network} \label{MRN}

The MS$^2$TAN (\underline{M}ulti-\underline{S}cale \underline{M}asked \underline{S}patial-\underline{T}emporal \underline{A}ttention \underline{N}etwork) is a multi-scale restoration network composed of $S$ residual-connected restoration modules and observed value replacement, as illustrated in Fig. \ref{fig:MRN}. Each module consists of Patch Embedding, MFE, and Patch Unembedding. These restoration modules can capture the spatiotemporal correlations of pixels at different scales to predict missing values and ultimately replace observed values as outputs.

\begin{figure}[h]
	\centering
	\includegraphics[width=\linewidth]{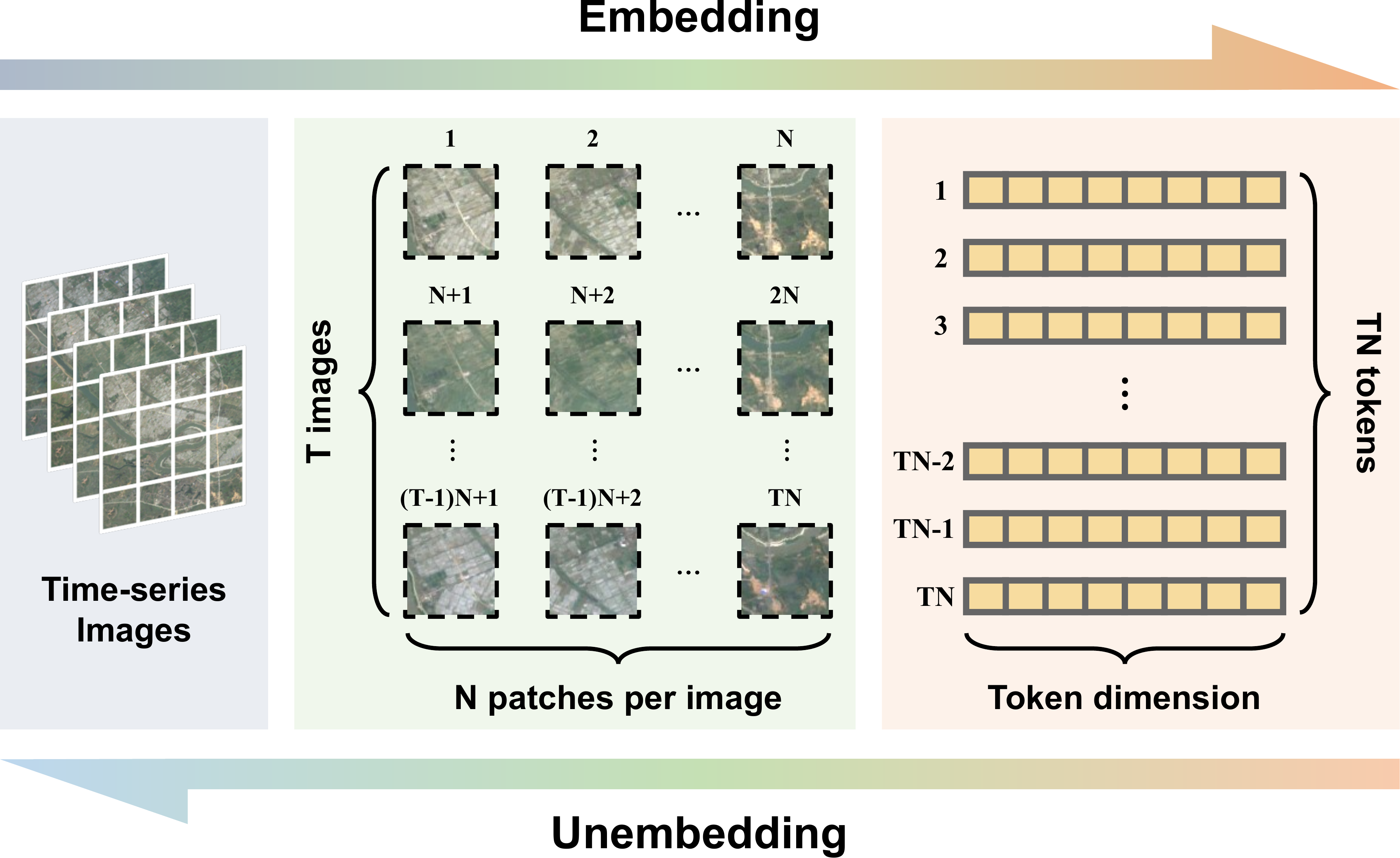}
	\caption{Illustration of Embedding and Unembedding processes. Embedding represents a series of transformations from left to right, and Unembedding represents the reverse.}
	\label{fig:embedding}
\end{figure}

\subsubsection{Embedding and Unembedding} \label{Emb-Unemb}

Following the approach of ViT, each image is partitioned into a series of patches, with the size of each patch determined by the hyperparameter \(P\). For each temporal image, \(N = HW/P^2\) patches are obtained. These patches are then aggregated and reorganized to form a patch sequence of length \(TN\), with each patch represented as a \(CP^2\)-dimensional vector. Equation \eqref{patchify} delineates the processing of the \(t\)-th temporal image \(X_\mathrm{in}^t\).
\begin{equation}
\begin{gathered}
A^t=\left[p_{1}^t,p_{2}^t,\cdots,p_{N}^t\right]=\operatorname{Patchify}(x_\mathrm{in}^t, P)\\
\text{where}\ x_\mathrm{in}^{t} \in \mathbb{R}^{C \times H \times W}\ \text{and}\ A^t \in \mathbb{R}^{N \times CP^2}
\end{gathered} \label{patchify}
\end{equation}

Next, a linear layer is utilized to project the vectors corresponding to each patch into a high-dimensional representation space. The images from different time steps are then aggregated to obtain the complete token sequence \(E\) :
%, as depicted in Eq. \eqref{patch2emb}.
\begin{equation}
\begin{gathered}
E^t = \mathrm{Linear}(A^t) \in \mathbb{R}^{N \times d_{\mathrm{emb}}}\\
E = \left[E^1,E^2,\cdots ,E^T\right] \in \mathbb{R}^{TN \times d_{\mathrm{emb}}}
\end{gathered} \label{patch2emb}
\end{equation}

The process of transforming $X_{\mathrm{in}}$ and $M$ step by step to obtain $E$ as described above is referred to as Embedding, which can be represented by Eq. \eqref{Embeding-Equal}. Conversely, if this operation is reversed, i.e., $E$ is restored to a patch sequence and then rearranged into an image, it is referred to as Unembedding, as shown in Eq. \eqref{Unembeding-Equal}. Fig. \ref{fig:embedding} illustrates the processes of Embedding and Unembedding on images.
\begin{gather}
E = \mathrm{Embedding}\left(X_{\mathrm{in}}, P\right) \label{Embeding-Equal}\\
X_{\mathrm{out}} = \mathrm{Unembedding}\left(E, P\right) \label{Unembeding-Equal}
\end{gather}

\subsubsection{MS$^2$TAN}

MS$^2$TAN consists of components such as Embedding, MFE, and Unembedding, where the patch size $P$ is an important hyperparameter, and the embedding scale of the $i$-th layer is denoted as $P^{\left(i\right)}$. The process of obtaining the embedding vector $\alpha$ from the input is described in Eq. \eqref{RRM-emb}. Subsequently, MFE is utilized to mine deep spatiotemporal correlation features $\beta$, as shown in Eq. \eqref{RRM-feature}. Finally, the features are decoded, unembedded back into image form, and added to the input, resulting in an intermediate result as depicted in Eq. \eqref{RRM-final}, where $\widetilde{Y}^{(i)}$ represents the $i$-th intermediate result.
\begin{gather}
\alpha = \mathrm{Embedding}\left(\widetilde{Y}^{(i-1)}, P^{\left(i\right)}\right) \label{RRM-emb}\\
\beta = \mathrm{MFE}\left(\alpha, M\right) \label{RRM-feature}\\
\widetilde{Y}^{(i)} = \mathrm{Unembedding}\left(\beta, P^{\left(i\right)}\right) + \widetilde{Y}^{(i-1)} \label{RRM-final}
\end{gather}

Let $\widetilde{Y}^{(0)}=X_{\mathrm{in}}$, and sequentially obtain $S$ intermediate results of the reconstruction. Eventually, the $S$-th result is taken as the output of MS$^2$TAN, as shown in Eq. \eqref{MRN-final}. 
\begin{gather}
\widetilde{Y} = \text{MS$^2$TAN}(X_{\mathrm{in}}, M) = \widetilde{Y}^{(S)}\label{MRN-final}
\end{gather}

\subsubsection{Observed Value Replacement}

For the original reconstruction results of this network, denoted as $\widetilde{Y}$, the observed values are replaced with their non-missing parts to obtain the final result $\widetilde{Y}_{\mathrm{out}}$, as shown in Eq. \eqref{replaceOut}.

\subsubsection{Multi-scale Restoration Strategy}
Common ViT-based models use a fixed patch size, which has been shown to be inefficient in dense prediction tasks like image segmentation and restoration \cite{wang2021pyramid}. In such tasks, smaller patch sizes often achieve better performance, but also come with greater computational costs. In order to balance the effect and efficiency, we apply a multi-scale restoration strategy, applying a larger patch size at a scale close to the input to obtain a coarse result, and applying a smaller patch size at a scale close to the output to obtain a fine result.

\subsection{Multi-Objective Joint Optimization} \label{MOO}
To fully exploit the reconstruction capability of the model, this paper proposes a multi-objective joint optimization method to train the network parameters. This method utilizes the ``Pixel-Structure-Perception'' Multi-Objective Loss Function to optimize the results generated by the model from the perspectives of structure, color, texture, shape, and spatial relationships, achieving high-quality image inpainting.

\subsubsection{Pixel-wise Loss}

Pixel-wise loss disregards the overall integrity of the image, treating the image as a collection of pixels, and comparing pixel by pixel to generate the image against the target image. It serves as the foundational loss for image reconstruction tasks. Equation \eqref{pixel-loss} illustrates its calculation process. The pixel-wise loss calculated here includes both missing and observed parts. The reduction of the loss in the observed part does not directly improve the performance of the model. However, it does smooth the model output, making the model training process more stable, and hence is also included.
\begin{gather} {\mathcal{L}}_{\textrm{pixel-wise}}=\frac{1}{TCHW}\left|\left|\eta-y\right|\right|_2^2\label{pixel-loss}
\end{gather}

\subsubsection{Structural Loss}

Pixel-wise loss is commonly employed in various time-series and visual tasks but fails to consider the correlation between pixels. Structural loss uses structural similarity \cite{SSIM} to measure the difference from the target image to optimise the visual consistency of the reconstruction results in terms of structure, contrast and luminance, and has been shown to have better performance in image reconstruction tasks \cite{zhao2015loss}. Equation \eqref{struct-loss} specifies the computation process of structural loss, where $\mu_\eta$ and $\mu_y$ denote the means of $\eta$ and $y$ respectively, $\sigma_\eta$ and $\sigma_y$ denote the variances of $\eta$ and $y$ respectively, $\sigma_{\eta y}$ denotes the covariance between $\eta$ and $y$, and $C_1$ and $C_2$ are constants.
\begin{equation}
\begin{aligned} {\mathcal{L}}_{\mathrm{structural}}&=1-\mathrm{SSIM}\left(\eta,y\right)\\
&=1-\frac{\left(2{\mu_\eta}{\mu_y}+C_1\right)\left(2\sigma_{\eta y}+C_2\right)}{\left(\mu_\eta^2+\mu_y^2+C_1\right)\left(\sigma_\eta^2+\sigma_y^2+C_2\right)}
\end{aligned}\label{struct-loss}
\end{equation}

\subsubsection{Perceptual Loss}

Perceptual loss \cite{johnson2016perceptual} is widely used in tasks like super-resolution and image generation, where pre-trained CNNs are used as the feature network to extract perceptual features such as texture details, image style, etc. Specifically, we use a VGG-16 \cite{VGG} with the fully connected layer removed as the feature network and the L2 loss of the feature map differences as the final loss. The calculation of perceptual loss is shown in Eq. \eqref{pcpt-loss}, where $\psi\left(\right)$ denotes the feature network and $d_\text{f}$ denotes the feature dimension.
\begin{equation}
\begin{aligned} {\mathcal{L}}_\mathrm{perceptual}=\frac{1}{d_\text{f}}\left|\left|\psi\left(\eta\right)-\psi\left(y\right)\right|\right|_2^2
\end{aligned}\label{pcpt-loss}
\end{equation}

\subsubsection{Multi-Objective Loss Function}

For each scale of MS$^2$TAN, the expression of the loss function $\mathcal{L}^{\left(i\right)}$ corresponding to its intermediate output $\widetilde{Y}^{(i)}$ is shown in Eq. \eqref{layer-loss}, where $\lambda_1$, $\lambda_2$ and $\lambda_3$ denote the weights of the pixel-wise loss, structural loss and perceptual loss respectively. The multi-objective loss function $\mathcal{L}$ considers the intermediate results of each scale, using their mean as the overall loss, as shown in Eq. \eqref{all-loss}.
\begin{gather}
\mathcal{L}^{\left(i\right)}\left(\eta, y\right)={\lambda_1}\mathcal{L}_{\text{pixel-wise}}+{\lambda_2}\mathcal{L}_{\mathrm{struct}}+{\lambda_3}\mathcal{L}_{\mathrm{perceptual}}\label{layer-loss}\\
\mathcal{L}=\frac{1}{S}\sum_{i=1}^{S}\mathcal{L}^{\left(i\right)}\left(\widetilde{Y}^{(i)}, Y\right)\label{all-loss}
\end{gather}

\section{Experiments} \label{Experiments}

\subsection{Settings}

To assess the performance of the proposed model in different scenarios, we conducted quantitative and qualitative experiments on two datasets and compared it with some mainstream methods. In addition, we conduct ablation experiments on the key innovations and discuss the balance between effectiveness and efficiency for models with different sizes.

\subsubsection{Compared Algorithms}

Based on the type of inputs and outputs, common image restoration methods can be categorized into three main types: single-input-single-output (SISO), dual-input-single-output (DISO), and multi-input-multi-output (MIMO). The SISO algorithms are the simplest type, but they struggle to achieve excellent performance, so we do not compare this category with ours. The DISO algorithm takes the target image and a temporal reference image as input and synthesizes a single reconstructed image. For these methods, we use the less missing temporal neighbor of the target image as the reference. The DISO methods we compare include simple replacement (replace missing values by copying data from reference image), LLHM \cite{LLHM}, WLR \cite{zeng2013recovering}, and STS-CNN \cite{STS-CNN}. The MIMO method utilizes a multi-temporal image containing missing data as input and outputs a reconstructed multi-temporal image, allowing the model to fully take into account temporal semantics. Compared methods of this type include last padding (Last), nearest neighbor padding (Nearest), and linear interpolation (Linear), which are commonly used in cartography \cite{inglada2015assessment}, as well as the U-TILISE \cite{stucker2023u}.

\subsubsection{Evaluation Metrics}

Through the sliding windows approach, we reconstructed and assessed data gaps across all time steps to facilitate a more comprehensive analysis. The quantitative experiments utilized mean absolute error (MAE), spectral angle mapper (SAM) \cite{kruse1993spectral}, mean peak signal-to-noise ratio (PSNR), and mean structural similarity (SSIM) \cite{SSIM} as evaluation metrics. In addition, for real data gaps where ground truth is missing, we clearly show the comparison of restoration results for qualitative assessment.

\subsubsection{Implement Details}
For MS$^2$TAN, we utilized the Adam optimizer with parameters ($\beta_1 = 0.9$, $\beta_2 = 0.999$) for training, and set the batch size to 8. The initial learning rate was set to $4 \times 10^{-4}$, with a decay schedule of every 100 epochs, and an early stopping strategy of 30 epochs on the validation set to prevent overfitting. For the other models we compared, we conducted our experiments while following the original authors' training settings and environment. The proposed MS$^2$TAN was implemented with PyTorch 1.12 framework and trained with a NVIDIA GeForce RTX 3090 24GB GPU on a Ubuntu 20.04 environment. Additionally, the test code of MS$^2$TAN is openly available on the Github at \href{https://github.com/CUG-BEODL/MS2TAN}{https://github.com/CUG-BEODL/MS2TAN}.

\subsection{Datasets}
We conducted experiments on two datasets, Landsat and EarthNet2021 \cite{Requena_2021_CVPR_Workshops}. Their details are described in Table \ref{tab:dataset-info}.

\begin{table*}[t]
\centering
\tabcolsep=1.4em
\renewcommand\arraystretch{1.35}
\caption{Quantitative Evaluation Results under \textbf{Dual-Temporal} Inputs Compared to the \textbf{DISO} Methods, using the Landsat and EarthNet2021 datasets. The results for the best-performing method for each metric are highlighted in \textbf{bold}.}
\begin{tabular}{c|c|cccc|cccc}
    \Xhline{2pt}
    \multirow{2}{*}{Method} & \multirow{2}{*}{Type} & \multicolumn{4}{c|}{\textbf{Landsat}} & \multicolumn{4}{c}{\textbf{EarthNet2021}} \\
    \cline{3-10}
    & & $\downarrow$ MAE & $\downarrow$ SAM & $\uparrow$ PSNR & $\uparrow$ SSIM & $\downarrow$ MAE & $\downarrow$ SAM & $\uparrow$ PSNR & $\uparrow$ SSIM \\
    \Xhline{1pt}

    Replace & \multirow{4}{*}{DISO} & 0.0210 & 2.81 & 32.67 & 0.8913 & 0.0146 & 3.10 & 34.04 & 0.9452 \\
    LLHM \cite{LLHM} & & 0.0161 & 1.95 & 34.64 & 0.9104 & 0.0121 & 2.59 & 35.54 & 0.9566 \\
    WLR \cite{zeng2013recovering} & & 0.0110 & 1.33 & 36.88 & 0.9312 & 0.0113 & 2.41 & 36.08 & 0.9605 \\
    STS-CNN \cite{STS-CNN} & & 0.0091 & 1.12 & 37.79 & 0.9423 & 0.0102 & 2.18 & 37.07 & 0.9647 \\
    \Xhline{0.4pt}
    \cellcolor{Lavender}{\textbf{MS$^2$TAN (Ours)}} & \cellcolor{Lavender}{DIDO} & \cellcolor{Lavender}{\textbf{0.0081}} & \cellcolor{Lavender}{\textbf{1.02}} & \cellcolor{Lavender}{\textbf{38.54}} & \cellcolor{Lavender}{\textbf{0.9503}} & \cellcolor{Lavender}{\textbf{0.0095}} & \cellcolor{Lavender}{\textbf{2.04}} & \cellcolor{Lavender}{\textbf{37.60}} & \cellcolor{Lavender}{\textbf{0.9671}} \\
    \Xhline{2pt}
\end{tabular}
\label{tab:diso_result}
\end{table*}

\begin{figure*}[t!]
\centering
\includegraphics[width=\linewidth]{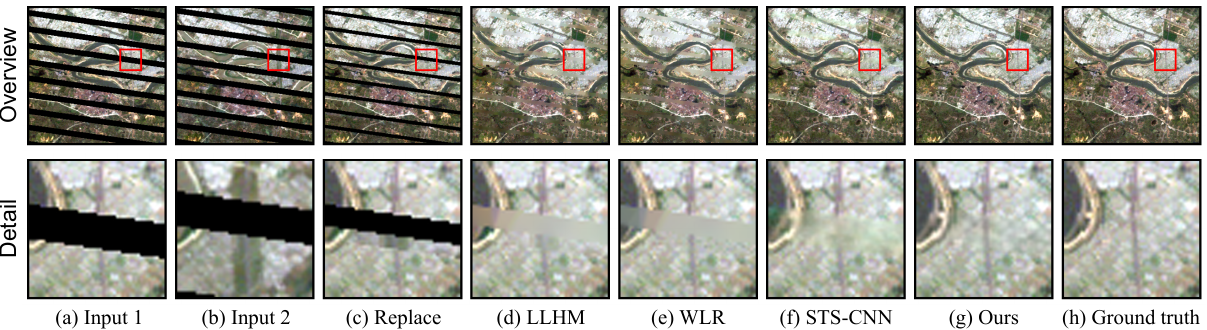}
\caption{Visual comparison results between MS$^2$TAN and the DISO algorithms on the Landsat dataset, with dead pixels shown in black. Row 2 details the highlighted area in row 1. From left to right: (a) Input 1 (target). (b) Input 2 (reference). (c)-(g) restoration results of each method. (h) Ground truth.}
\label{fig:diso_visual}
\end{figure*}

\subsubsection{Landsat}

The Landsat dataset contains a large number of Top-of-Atmosphere (TOA) time-series captured by Landsat-5/7/8 satellites, with approximately uniform time intervals. The original images are cropped into small patches of size $120 \times 120$ pixels with a spatial resolution of 30 meters. Each image includes 6 bands (Blue, Green, Red, NIR, SWIR1, and SWIR2), as well as the SLC-off or cloud mask provided in the QA band. In terms of spatial extent, we focus on sequences from multiple study regions in China and the United States.

\subsubsection{EarthNet2021}

The EarthNet2021 \cite{Requena_2021_CVPR_Workshops} dataset comprises over 28,000 Sentinel-2 TOA time-series observations from the European continent, with uniformly spaced time intervals. Each image consists of 4 bands (Blue, Green, Red and NIR), at a size of $128 \times 128$ pixels, downsampled to a spatial resolution of 20-meters, accompanied by cloud probability maps. We utilized 80\% of sequences from the training split for training purposes, with the remaining set aside for validation. For testing, we used sequences from the \textit{iid} test split.

\subsubsection{Preprocessing}

We follow the preprocessing protocol of EarthNet2021 to crop the values of the optical images to the range $\left[0, 10~000\right]$ and then normalize them to the unit range $\left[0, 1\right]$. To match the multi-scale patch division strategy of MS$^2$TAN, we resize the input images to a common multiple of the patch size of the MSTA at different scales. In our experiments, we use three scales (patch size is 8, 10, and 12 respectively) which have a least common multiple of 120. We therefore centrally cropped the images in the EarthNet2021 dataset to $120 \times 120$ pixels, in line with the Landsat dataset.

\subsection{Results}

\begin{table*}[t]
\centering
\tabcolsep=1.4em
\renewcommand\arraystretch{1.35}
\caption{Quantitative Evaluation Results under \textbf{Multit-Temporal} Inputs Compared to the \textbf{MIMO} Methods, using the Landsat and EarthNet2021 datasets. The results for the best-performing method for each metric are highlighted in \textbf{bold}.}
\begin{tabular}{c|c|cccc|cccc}
    \Xhline{2pt}
    \multirow{2}{*}{Method} & \multirow{2}{*}{Type} & \multicolumn{4}{c|}{\textbf{Landsat}} & \multicolumn{4}{c}{\textbf{EarthNet2021}} \\
    \cline{3-10}
    & & $\downarrow$ MAE & $\downarrow$ SAM & $\uparrow$ PSNR & $\uparrow$ SSIM & $\downarrow$ MAE & $\downarrow$ SAM & $\uparrow$ PSNR & $\uparrow$ SSIM \\
    \Xhline{1pt}
    
    Last & \multirow{5}{*}{MIMO} & 0.0223 & 3.14 & 32.11 & 0.8862 & 0.0148 & 3.17 & 33.68 & 0.9439 \\
    Nearest & & 0.0198 & 2.54 & 33.28 & 0.8957 & 0.0128 & 2.74 & 35.04 & 0.9534 \\
    Linear & & 0.0103 & 1.26 & 37.20 & 0.9348 & 0.0110 & 2.35 & 36.48 & 0.9620 \\
    U-TILISE \cite{stucker2023u} & & 0.0082 & 1.05 & 38.45 & 0.9491 & 0.0086 & 1.87 & 38.29 & 0.9702 \\
    \Xcline{1-1}{0.4pt}\cline{3-10}
    \cellcolor{Lavender}{\textbf{MS$^2$TAN (Ours)}} & \cellcolor{Lavender} & \cellcolor{Lavender}{\textbf{0.0074}} & \cellcolor{Lavender}{\textbf{0.96}} & \cellcolor{Lavender}{\textbf{39.01}} & \cellcolor{Lavender}{\textbf{0.9552}} & \cellcolor{Lavender}{\textbf{0.0078}} & \cellcolor{Lavender}{\textbf{1.71}} & \cellcolor{Lavender}{\textbf{38.91}} & \cellcolor{Lavender}{\textbf{0.9728}} \\
    \Xhline{2pt}
\end{tabular}
\label{tab:mimo_result}
\end{table*}

\begin{figure*}[t!]
\centering
\includegraphics[width=\linewidth]{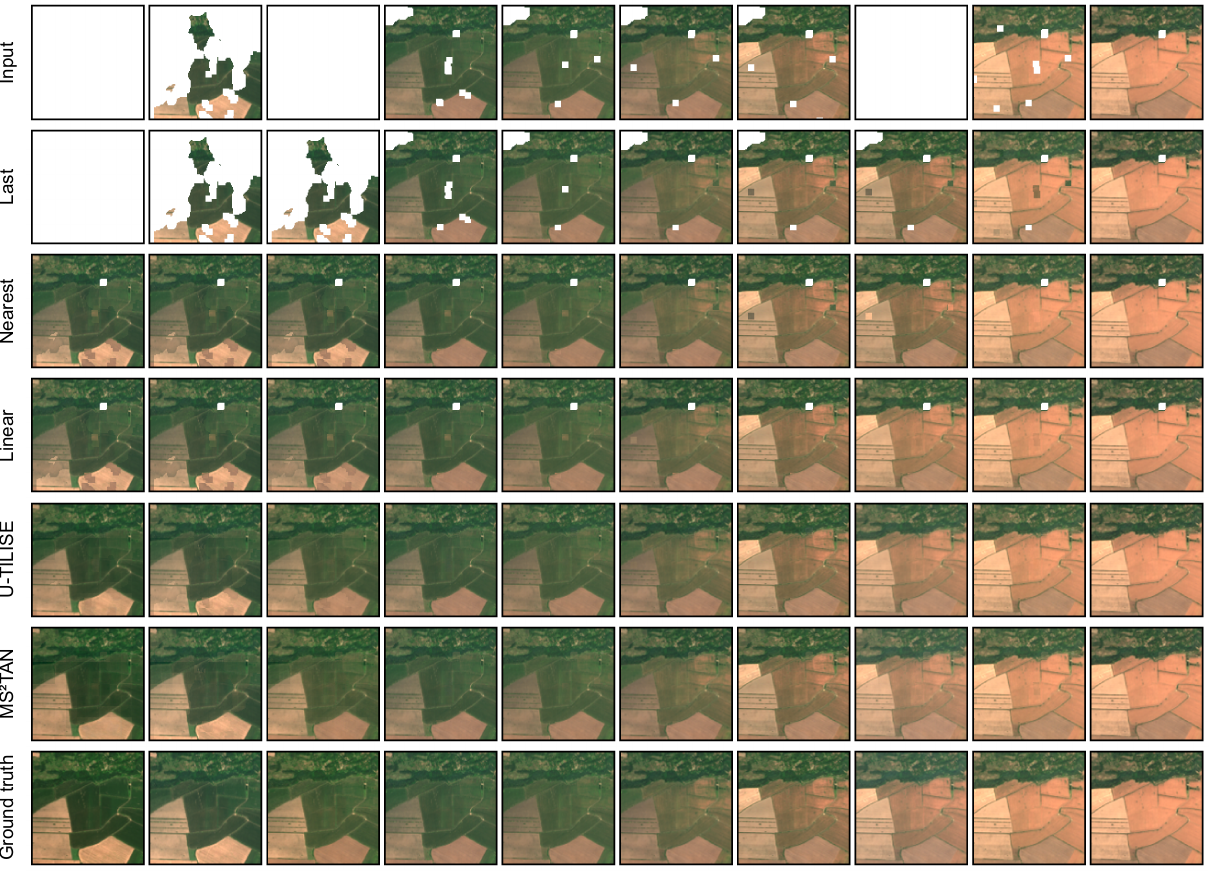}
\caption{Visual comparison results between MS$^2$TAN and the MIMO algorithms on the EarthNet2021 dataset, with thick cloud shown in white. Each row shows a time-ordered image sequence. From top to bottom: input sequence (row 1), restoration results of each method (rows 2-6), and ground truth (row 7).}
\label{fig:mimo_visual}
\end{figure*}

To quantitatively compare restoration results, we evaluated our proposed approach as well as mainstream methods on synthetic data gaps. Specifically, we performed comparison experiments under two different conditions: dual-temporal inputs (vs. the DISO algorithms), and multi-temporal inputs (vs. the MIMO algorithms). The quantitative evaluation results on the Landsat and EarthNet2021 datasets are shown in Table \ref{tab:diso_result} and \ref{tab:mimo_result}, and visual eaxamples are illustrated in Figs. \ref{fig:diso_visual} and \ref{fig:mimo_visual}. Furthermore, Figs. \ref{fig:real_landsat} and \ref{fig:real_earthnet} shows the restoration results in real data gaps to validate the generalisation performance of MS$^2$TAN.

\subsubsection{Comparison under Dual-Temporal Inputs}

We conducted comparison experiments with the DISO algorithms using synthetic data gaps in the Landsat and EarthNet2021 datasets. As illustrated in Fig. \ref{fig:diso_visual}, (a) and (b) represent the two images with data gaps used as input, while (c)-(g) sequentially display the recovery results obtained from five methods, with (h) showing the ground truth of the target image. As depicted in (c)-(f), all compared methods to some extent produced discontinuous fine features. This is because the time-assisted images were unable to completely cover the missing regions, leading the LLHM, and WLR to rely on LRPM (Laplacian Prior Regularization Method \cite{zeng2013recovering}) to fill in the remaining gaps, resulting in very blurry noise bands in the reconstruction results. Although the end-to-end strategy-based STS-CNN model can repair these blanks, it fails to accurately restore the original features. In contrast, the proposed MS$^2$TAN, integrating a “Pixel-Structure-Perception” Multi-Objective Joint Optimization approach, achieves superior texture and structural detail recovery, as shown in (g). In the quantitative evaluation shown in Table \ref{tab:diso_result}, MS$^2$TAN outperforms other comparative methods across all evaluation metrics. Compared to the previous SOTA method STS-CNN, MS$^2$TAN exhibits a 10.99\%/6.86\% (in two datasets respectively) reduction in MAE and a 0.75 dB/0.53 dB increase in PSNR, showcasing its robust capability in utilizing spatiotemporal data. 

\subsubsection{Comparison under Multi-Temporal Inputs}

\begin{figure*}[t!]
\centering
\includegraphics[width=\linewidth]{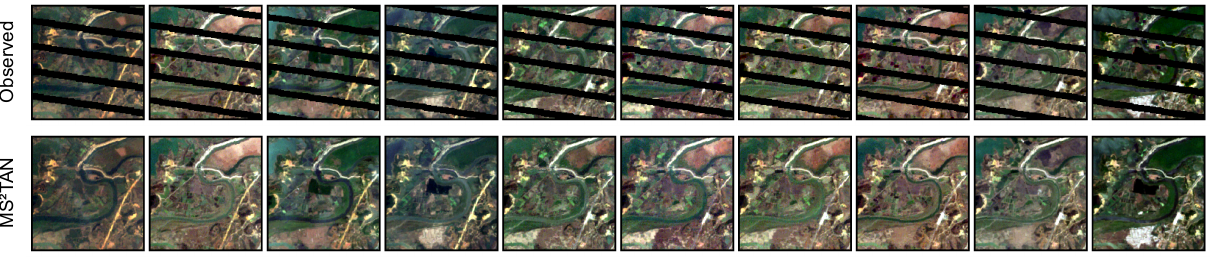}
\caption{Visual evaluation of restoration results for real ETM+ SLC-off images from the Landsat dataset, with dead pixels shown in black. The first and second rows show the observed sequence and the restoration results of the MS$^2$TAN, respectively.}
\label{fig:real_landsat}
\end{figure*}

\begin{figure*}[t!]
\centering
\includegraphics[width=\linewidth]{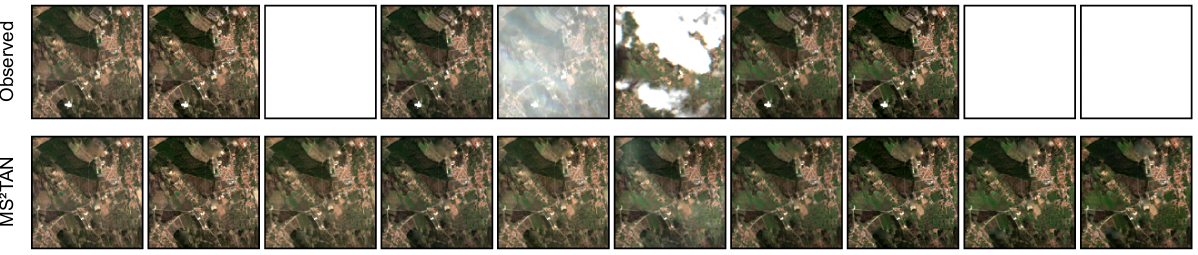}
\caption{Visual evaluation of restoration results for real Thick Cloud Cover images in the EarthNet2021 dataset, with clouds shown in white. The first and second rows show the observed sequence and the restoration results of the MS$^2$TAN, respectively.}
\label{fig:real_earthnet}
\end{figure*}

Similar to the dual-temporal input experiment, we conducted comparative experiments with the MIMO algorithms under multi-temporal input conditions. Visual comparative results are illustrated in Fig. \ref{fig:mimo_visual}, displaying, from top to bottom: the input image sequence, the restoration outcomes of five methods, and the ground truth. Since traditional linear methods rely on time-series information, they cannot restore regions in the input sequence where all time steps are missing. In contrast, U-TILISE can recover data gaps of any shape, but still exhibits significant color shifts in irregularly shaped gaps, reducing the quality of the results, as shown in the first three columns of Fig. \ref{fig:mimo_visual}. Benefiting from the Masked Spatial-Temporal Attention mechanism, MS$^2$TAN can effectively handle the edges of missing regions, achieving more harmonious seams. Notably, the Multi-Objective Optimization makes MS$^2$TAN perform better in restoring structural and textural details in the missing regions. In some extreme cases (e.g., the upper-left region of the first 8 columns), due to a lack of sufficient information, MS$^2$TAN may not accurately reproduce the actual changes but can still output coherent images with natural color transitions. Compared to the U-TILISE, MS$^2$TAN exhibits a 9.76\%/9.30\% reduction in MAE, a 0.56 dB/0.62 dB increase in PSNR, and achieves higher visual consistency.

\subsubsection{Restoration of the Real Data Gaps}

To evaluate the generalization performance of the proposed MS$^2$TAN, we applied pre-trained model weights to real ETM+ SLC-off sequences and thick cloud-covered sequences. Due to the lack of ground truth reflectance data, the assessment was conducted based solely on visual inspection. Visualization of the restoration results is provided in Figs. \ref{fig:real_landsat} and \ref{fig:real_earthnet}. In the restored ETM+ SLC-off images, the roads and rivers, which are divided by striped missing areas, are rationally connected in the restoration results with the original structure maintained. Obvious ground surface changes brought by long time intervals are also taken into account. MS$^2$TAN utilizes Masked Spatial-Temporal Attention to learn the distribution of missing data, thereby capturing temporal evolution patterns from neighboring frames and fitting spatial textures, achieving high-precision data restoration. For the thick cloud-covered images, the method accurately restored all missing data, including both large-scale and fragmented gaps. Based on the assistance of long time-series, continuous large-area data gaps can also be repaired reasonably well and keep the original evolutionary trend. Irregularly shaped data gaps (e.g., the images in the sixth column in Fig. \ref{fig:real_earthnet}) are repaired to blend in with the spatial domain, preserving high textural and structural coherence. By applying Masked Spatial-Temporal Attention and Multi-Objective Joint Optimization, MS$^2$TAN exhibits excellent performance in handling complex and large-scale data gaps, which often present challenges in traditional data recovery methods, as evident in the detailed visual evaluation.

\subsection{Validation Studies}

To validate the effectiveness of the current architecture, we experimented with some variants of MS$^2$TAN and obtained quantitative results as shown in Table \ref{tab:variants}. We discuss each of the mechanisms used below.

\begin{table*}[htbp]
  \centering
  \renewcommand\arraystretch{1.35}
  \setlength{\tabcolsep}{0.6em}
  \caption{Quantitative Evaluation of Different \textbf{MS$^2$TAN Variants} that derived from whether Separated Spatial-Temporal Attention (\textbf{Sep. Attn}), Attention Mask Mechanism (\textbf{Attn Mask}), and Multi-scale Restoration Strategy (\textbf{Multi-Scale}) are enabled. Statistics information and evaluation metrics are provided on the right. All experiments are performed under Multi-Temporal Inputs.}
  \begin{tabular}{ccc|cc|cccc|cccc}
      \Xhline{2pt}
      \multicolumn{3}{c|}{\textbf{Features}} & \multicolumn{2}{c|}{\textbf{Statistics}} & \multicolumn{4}{c|}{\textbf{Landsat}} & \multicolumn{4}{c}{\textbf{EarthNet2021}} \\
      \hline
      Sep. Attn & Attn Mask & Multi-Scale & Params (M) & FLOPs (G) & $\downarrow$ MAE & $\downarrow$ SAM & $\uparrow$ PSNR & $\uparrow$ SSIM & $\downarrow$ MAE & $\downarrow$ SAM & $\uparrow$ PSNR & $\uparrow$ SSIM \\
      \Xhline{1pt}
      \checkmark & \checkmark & \checkmark & \textbf{4.33} & \textbf{8.756} & 0.0074 & 0.96 & 39.01 & 0.9552 & \textbf{0.0078} & \textbf{1.71} & \textbf{38.91} & \textbf{0.9728} \\
      \Xhline{0.4pt}
      & \checkmark & \checkmark & \textbf{4.33} & 14.626 & \textbf{0.0073} & \textbf{0.95} & \textbf{39.08} & \textbf{0.9560} & 0.0083 & 1.81 & 38.53 & 0.9714 \\
      \checkmark & & \checkmark & \textbf{4.33} & \textbf{8.756} & 0.0096 & 1.18 & 37.47 & 0.9384 & 0.0112 & 2.38 & 36.37 & 0.9609 \\
      \checkmark & \checkmark & & 6.61 & 10.130 & 0.0086 & 1.07 & 38.09 & 0.9459 & 0.0095 & 2.05 & 37.60 & 0.9672 \\
%      \Xhline{0.4pt}
      & & & 6.61 & 16.473 & 0.0101 & 1.25 & 37.24 & 0.9356 & 0.0127 & 2.73 & 35.19 & 0.9538 \\
      \Xhline{2pt}
  \end{tabular}
  \label{tab:variants}
\end{table*}

\subsubsection{Separated Spatial-Temporal Attention}

The original ViT, after embedding images into token sequences, feeds the entire sequence into a multi-head self-attention module. However, the computational complexity of self-attention is proportional to the square of the sequence length. For a sequence of remote sensing images ($t$ images, $n$ patches per image), the computational complexity is as high as $O\left(t^2 \times n^2\right)$. Whereas, separating temporal attention from spatial attention and computing them sequentially reduces the complexity to $O\left(t^2+ n^2\right)$. Although the perceptual field of the separated spatial-temporal attention is reduced in a single operation, we alternate multiple temporal and spatial attention modules to spread the perceptual field to the whole spatiotemporal range. Meanwhile, the extension of the perceptual field along the temporal axis and the spatial plane matches the characteristics of the time-series remote sensing images, which have small displacement in time and large spatial correlation, and can fully integrate the spatial and temporal information. As shown in the first two rows of Table \ref{tab:variants}, the use of separated spatial-temporal attention reduces the computational effort by 40.1\% with almost the same accuracy in the Landsat dataset, while the accuracy in EarthNet2021 is slightly improved.

\begin{figure}[t]
\centering
\includegraphics[width=\linewidth]{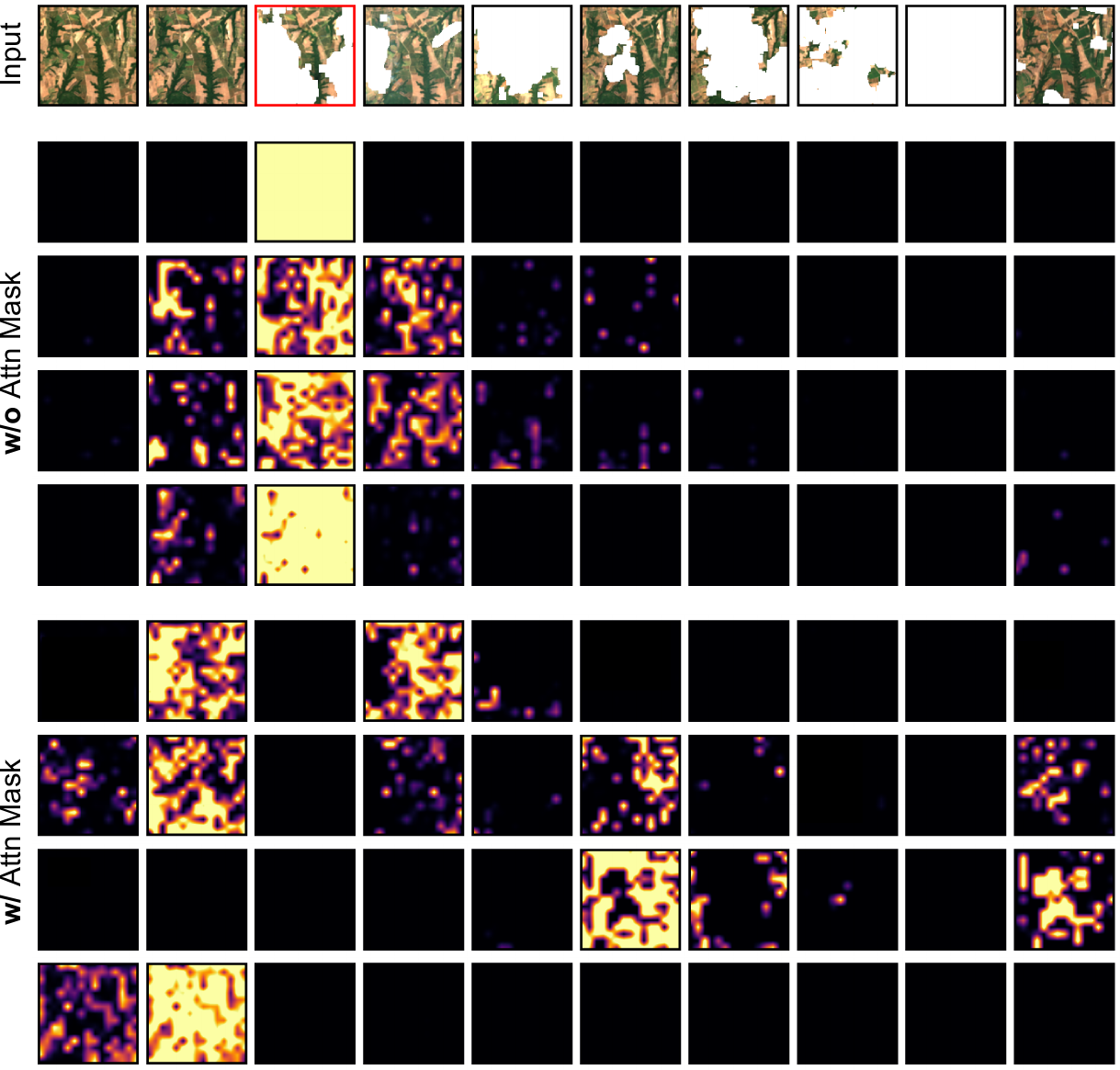}
\caption{Visualization of Temporal Attention Distribution in the first MSTA unit. We present the attention scores for repairing the 3rd frame (highlighted with a red box in row 1) of the sequence. It is observed that, after incorporating the mask, the temporal attention is no longer restricted to the vicinity of the target time but instead focuses on a broader range of valid pixels. From top to bottom: Input (row 1), \textbf{w/o} Mask (rows 2-5), and \textbf{w/} Mask (rows 6-9).}
\label{fig:time_attn_map}
\end{figure}

\begin{figure}[t]
\centering
\includegraphics[width=\linewidth]{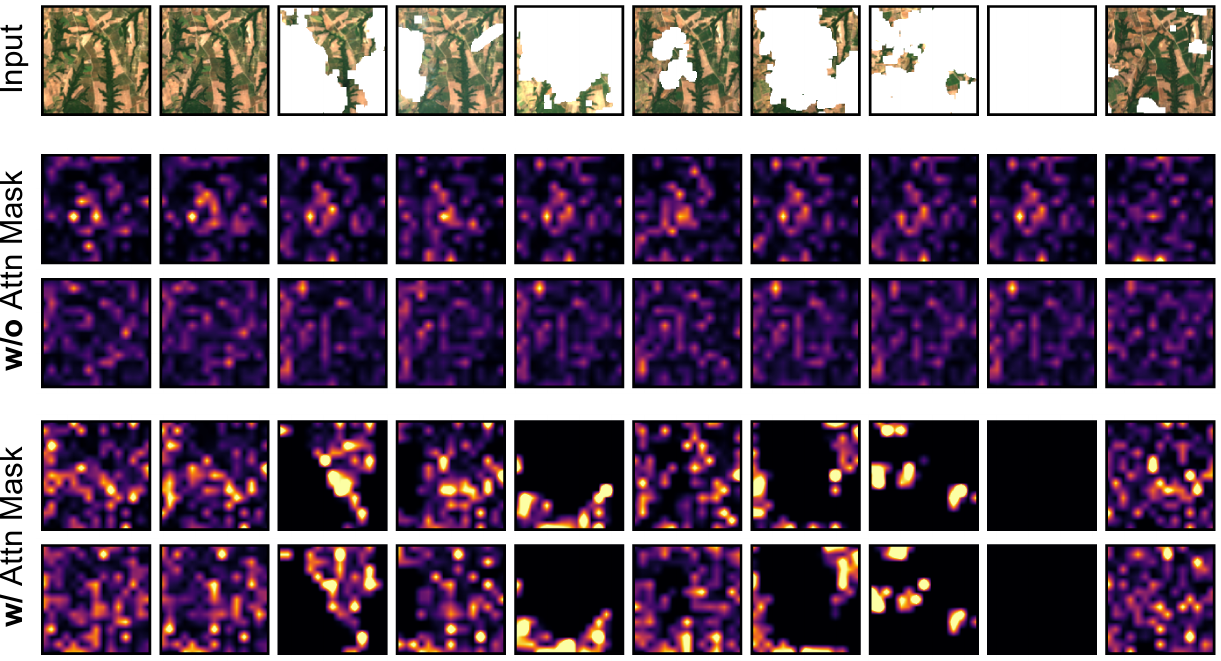}
\caption{Visualisation of Spatial Attention Distribution in the first MSTA unit. We show the average attention scores for imputing each patch. With the mask, spatial attention focuses on the edges of the valid pixels, extracting textural and structural features, rather than spreading around as it would be without the mask. From top to bottom: Input (row 1), \textbf{w/o} Mask (rows 2-3), and \textbf{w/} Mask (rows 4-5).}
\label{fig:space_attn_map}
\end{figure}

\begin{figure}[t]
\centering
\includegraphics[width=\linewidth]{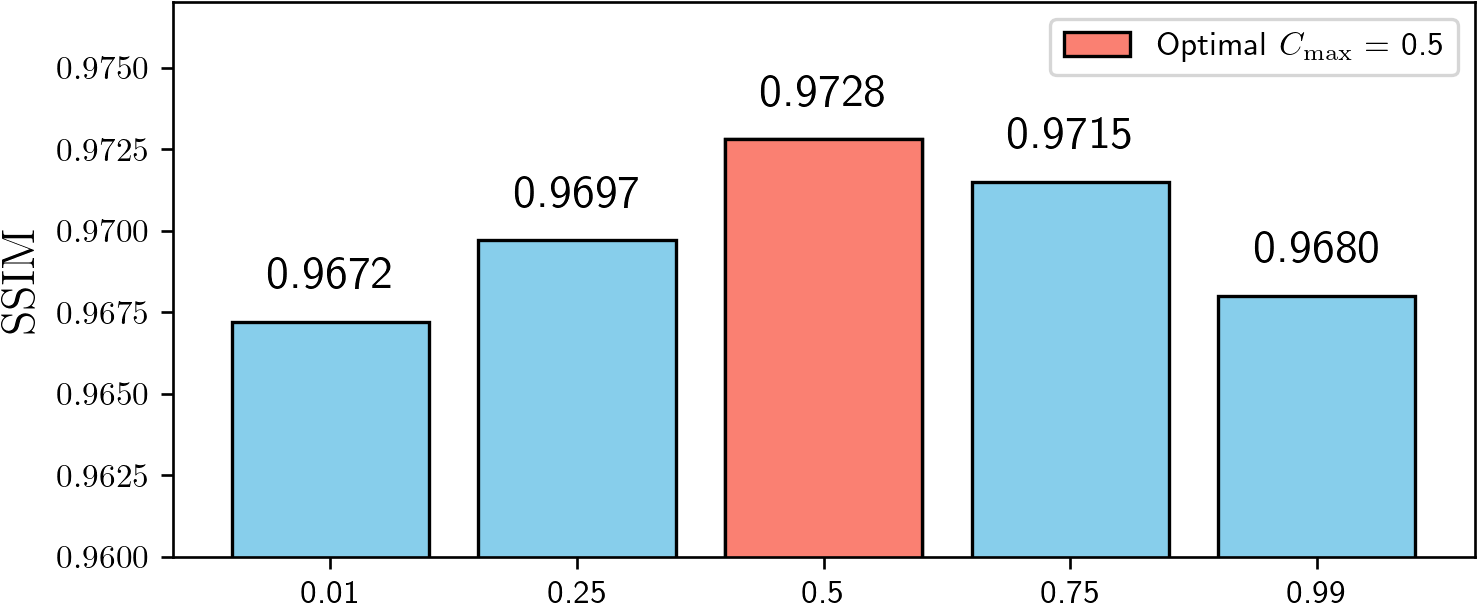}
\caption{Ablation Experiment Results for Optimal $C_\text{max}$ Value in MSTA. The horizontal axis represents different $C_\text{max}$ values, and the vertical axis represents the SSIM in the EarthNet2021 dataset.}
\label{fig:ablation_C_max}
\end{figure}

\subsubsection{Attention Mask in MSTA}

To verify the effectiveness of the attention mask in MSTA, we visualized the distribution of attention. Fig. \ref{fig:time_attn_map} shows the distribution of temporal attention with and without the attention mask, while Fig. \ref{fig:space_attn_map} illustrates the spatial attention distribution. Without the mask, certain attention heads focus excessively on patches with missing data, while others focus excessively on their own patch, providing no additional information to impute missing values and leading to unnecessary computational costs. With the mask mechanism in place, the attention heads no longer focus on missing-data patches or themselves, but instead extract information from spatiotemporal domains, obtaining a larger effective receptive field to achieve higher Imputation accuracy. Within the MSTA framework, the maximum allowable missing rate, $C_\text{max}$, is a crucial hyperparameter. Setting it too high can cause patches with missing data to distort token distribution, while a too-low value may result in information loss. Consequently, we conducted ablation experiments on different values of $C_\text{max}$, with the results shown in Fig. \ref{fig:ablation_C_max}. The findings indicate that an optimal $C_\text{max}$ yields the best reconstruction accuracy, and thus $C_\text{max}$ is set to 0.5 for subsequent experiments in this paper.

\subsubsection{Multi-scale Restoration}

\begin{figure}[t]
\centering
\includegraphics[width=\linewidth]{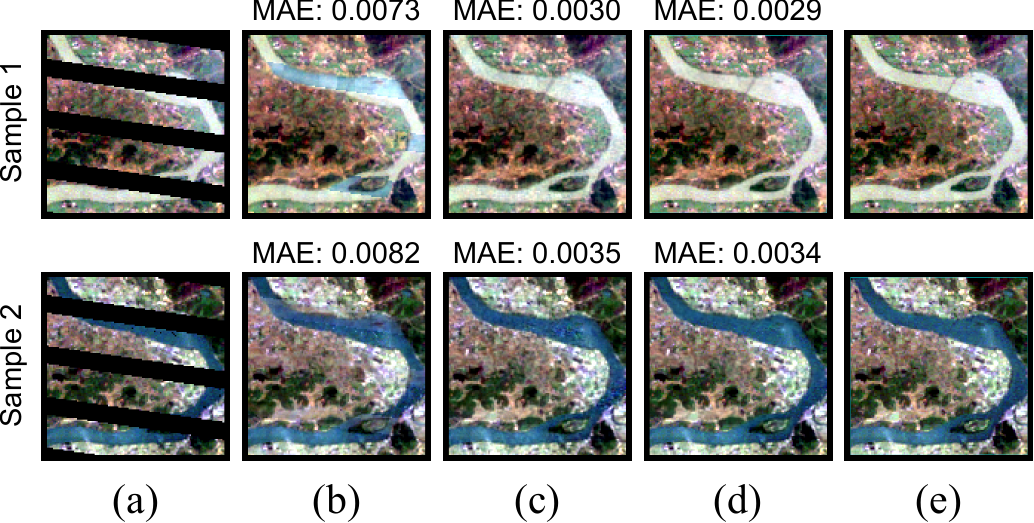}
\caption{Visualisation of intermediate results for multi-scale restoration. The values above indicate the MAE compared to the ground truth. (a) Input. (b)-(d) Intermediate outputs at each scale. (e) Ground truth.}
\label{fig:multi_scale_restoration}
\end{figure}

MS$^2$TAN employs a multi-scale restoration strategy to refine the restoration results layer by layer. To verify its effectiveness, we take the SLC-off reconstruction task as an example and compare the intermediate outputs of each layer in Fig. \ref{fig:multi_scale_restoration}. With the gradual deepening of the restoration stages, the restoration results are gradually refined to better capture the details and structural features in the image. Specifically, the coarse repair stage addresses global textures and structural issues, while the fine repair stage further optimizes local details and spectral consistency. However, despite the effectiveness of the multi-scale approach, introducing too many scale levels leads to diminishing returns, with further improvements becoming negligible or even introducing overfitting and increased computational costs. Therefore, selecting an appropriate number of scale levels is crucial.

\subsubsection{Multi-Objective Joint Optimization}

\begin{table}[t]
\centering
\tabcolsep=0.9em
\renewcommand\arraystretch{1.35}
\caption{Ablation Analysis for \textbf{Multi-Objective Joint Optimization}, where $\lambda_1$, $\lambda_2$ and $\lambda_3$ denote the weights of the pixel-wise loss, structural loss and perceptual loss respectively.}
\begin{tabular}{ccc|cccc}
	\Xhline{2pt}
	\multicolumn{3}{c|}{\textbf{Loss weights}} & \multicolumn{4}{c}{\textbf{EarthNet2021}} \\
	\Xhline{0.4pt}
	$\lambda_1$ & $\lambda_2$ & $\lambda_3$ & $\downarrow$ MAE & $\downarrow$ SAM & $\uparrow$ PSNR & $\uparrow$ SSIM \\
	\Xhline{1pt}
	1 & - & - & 0.0120 & 2.55 & 35.76 & 0.9572 \\
	0.9 & 0.1 & - & 0.0091 & 1.97 & 37.94 & 0.9686 \\
	0.9 & - & 0.1 & 0.0102 & 2.20 & 37.09 & 0.9645 \\
	- & 0.5 & 0.5 & 0.0210 & 4.43 & 31.72 & 0.9241 \\
	\Xhline{0.4pt}
	0.9 & 0.05 & 0.05 & \textbf{0.0078} & \textbf{1.71} & \textbf{38.91} & \textbf{0.9728} \\
	\Xhline{2pt}
\end{tabular}
\label{tab:multi-objective}%
\end{table}%

In Multi-Objective Joint Optimization, the weights assigned to each loss function are critical. To investigate this, we conducted an ablation study on the allocation of loss weights. Table \ref{tab:multi-objective} presents different weight configurations for pixel-wise loss ($\lambda_1$), structural loss ($\lambda_2$), and perceptual loss ($\lambda_3$), along with their corresponding results on the EarthNet2021 dataset. The findings show that combining pixel-wise loss with either structural or perceptual loss leads to significant performance improvements (cf. rows 2-3), with the best results achieved when both are combined. However, when only structural and perceptual losses are applied, the model struggles to converge, resulting in poor performance (cf. row 4). Therefore, pixel-wise loss plays a dominant role in multi-objective optimization, and its combination with structural and perceptual losses is essential for achieving high restoration accuracy and visual consistency.

\begin{table*}[htbp]
  \centering
  \tabcolsep=0.5em
  \renewcommand\arraystretch{1.5}
  \caption{Hyperparameter settings and Evaluation of MS$^2$TAN with different sizes. The hyperparameters include the restoration scales number ($S$), patch size ($P$), embedding dimensions ($d_\text{emb}$), attention heads number ($h$), dimensions of vectors Q, K and V in self-attention ($d_\text{qkv}$), and the MSTA units number ($L$), where bracketed lists indicate the hyperparameters for each scale.}
  \begin{tabular}{c|cccccc|cc|cccc}
    \Xhline{2pt}
    \multirow{2}{*}{\textbf{Model}} & \multicolumn{6}{c|}{\textbf{Hyperparameters}} & \multicolumn{2}{c|}{\textbf{Statistics}} & \multicolumn{4}{c}{\textbf{EarthNet2021}} \\
    \cline{2-13}
    & $S$ & $P$ & $d_\text{emb}$ & $h$ & $d_\text{qkv}$ & $L$ & Params (M) & FLOPs (G) & $\uparrow$ PSNR & $\uparrow$ SSIM & $\uparrow$ PSNR & $\uparrow$ SSIM\textbf{} \\
    \Xhline{1pt}
    MS$^2$TAN & 3 & (12, 10, 8) & (256, 192, 128) & (8, 6, 4) & (32, 32, 32) & (2, 2, 2) & 4.33 & 8.756 & 0.0078 & 1.71 & 38.91 & 0.9728 \\
    \Xhline{0.4pt}
    MS$^2$TAN-L & 3 & (12, 10, 8) & (384, 256, 192) & (8, 6, 4) & (48, 48, 48) & (4, 4, 4) & 17.04 & 34.363 & \textbf{0.0076} & \textbf{1.67} & \textbf{39.03} & \textbf{0.9733} \\
    MS$^2$TAN-S & 2 & (12, 10) & (192, 128) & (8, 6) & (24, 24) & (2, 2) & \textbf{2.07} & \textbf{3.567} & 0.0089 & 1.92 & 38.07 & 0.9692 \\
    \Xhline{2pt}
  \end{tabular}
  \label{tab:hyperparameters-analysis}
\end{table*}

\subsubsection{Balance between Effectiveness and Efficiency}

Scalability is a key strength of ViT-based models, and MS$^2$TAN is no exception. To strike a balance between effectiveness and efficiency, we adjusted several critical hyperparameters to obtain two MS$^2$TAN variants and evaluated them on EarthNet2021 dataset (see Table \ref{tab:hyperparameters-analysis}). The experimental results reveal that when the model is too small (MS$^2$TAN-S), performance significantly degrades; conversely, when the model is too large (MS$^2$TAN-L), there is only a slight performance gain, while the computational cost increases drastically, making the approach less efficient. Overall, the current configuration of parameters and computational load represents an optimal “sweet spot” where good performance is maintained without excessive computational overhead.

\section{Conclusion} \label{Conclusion}
This paper presents a novel approach to reconstruct time-series remote sensing images using the Multi-Scale Masked Spatial-Temporal Attention Network (MS$^2$TAN). MS$^2$TAN takes an image sequence with arbitrary data gaps as input and generates a clear and complete time-series. Compared with the existing methods, the proposed Masked Spatial-Temporal Attention (MSTA) mechanism improves the efficiency of spatiotemporal information utilization, reduces the high computational cost of the original ViT, and obtains higher restoration accuracy. In addition, the application of the Multi-scale Restoration strategy and Multi-Objective Joint Optimization further enhances the texture and structural consistency of the reconstructed images. MS$^2$TAN achieves higher imputation accuracy than the mainstream methods in quantitative experiments and exhibits better visual effects in eliminating real data gaps. Ablation experiments also confirm the contributions of primary innovations to the results.

Despite the effectiveness of the proposed method in addressing ETM+ SLC-off and thick cloud cover, there are some unavoidable limitations. For instance, the model needs to simulate various shapes of data gaps in the dataset to adequately learn to extract the features of the missing value distribution in order to achieve better restoration results, which leads to the model needing more time for training. Therefore, future research could incorporate network components pre-trained on large-scale datasets to improve model convergence speed and generalization performance.

\section*{Acknowledgments}

This work was supported in part by the National Natural Science Foundation of China (No. 42471505); in part by the National Key R\&D Program of China (Project No. 2022YFC3800700).

\appendix  % for no appendix heading

Table \ref{tab:dataset-info} details the statistics of the dataset we used, including key information such as data source, resolution, and number of sequences.

\begin{table}[htbp]
  \centering
  \tabcolsep=1em
  \renewcommand\arraystretch{1.5}
  \caption{Details of the Landsat and EarthNet2021 Datasets.}
    \begin{tabular}{cllcccc}
    \Xhline{2pt}
    \multicolumn{3}{c|}{}  & \multicolumn{2}{c}{\textbf{Landsat}} & \multicolumn{2}{c}{\textbf{EarthNet2021}} \\
    \Xhline{1pt}
    \multicolumn{3}{l|}{Data source} & \multicolumn{2}{c}{Landsat-5/7/8} & \multicolumn{2}{c}{Sentinel-2} \\
%    \Xhline{0.4pt}
    \multicolumn{3}{l|}{Restoration task} & \multicolumn{2}{c}{SLC-off \& Thick Cloud} & \multicolumn{2}{c}{Thick Cloud} \\
%    \Xhline{0.4pt}
    \multicolumn{3}{l|}{Spatial scope} & \multicolumn{2}{c}{China \& USA} & \multicolumn{2}{c}{Europe} \\
    \multicolumn{3}{l|}{Temporal scope} & \multicolumn{2}{c}{2001 - 2011} & \multicolumn{2}{c}{2016 - 2020} \\
    \Xhline{0.4pt}
    \multicolumn{3}{l|}{Spectral bands} & \multicolumn{2}{c}{6 bands} & \multicolumn{2}{c}{4 bands} \\
    \multicolumn{3}{l|}{Image size} & \multicolumn{2}{c}{120 $\times$ 120} & \multicolumn{2}{c}{128 $\times$ 128} \\
    \multicolumn{3}{l|}{Spatial Resolution} & \multicolumn{2}{c}{30-m} & \multicolumn{2}{c}{20-m} \\
%    \Xhline{0.4pt}
    \multicolumn{3}{l|}{Temporal length} & \multicolumn{2}{c}{10} & \multicolumn{2}{c}{30} \\
    \multicolumn{3}{l|}{Temporal interval} & \multicolumn{2}{c}{average 30 days} & \multicolumn{2}{c}{constant 5 days} \\
    \Xhline{0.4pt}
    \multicolumn{3}{l|}{Number of samples} & \multicolumn{2}{c}{11913} & \multicolumn{2}{c}{28123} \\
          & \multicolumn{2}{r|}{-- Train} & \multicolumn{2}{c}{7625} & \multicolumn{2}{c}{18955} \\
          & \multicolumn{2}{r|}{-- Valid} & \multicolumn{2}{c}{1905} & \multicolumn{2}{c}{4949} \\
          & \multicolumn{2}{r|}{-- Test} & \multicolumn{2}{c}{2383} & \multicolumn{2}{c}{4219} \\
    \Xhline{2pt}
    \end{tabular}%
  \label{tab:dataset-info}%
\end{table}%

\ifCLASSOPTIONcaptionsoff
  \newpage
\fi

\bibliographystyle{IEEEtran}
\bibliography{MyRefs}

\end{document}